\documentclass[conference]{IEEEtran}
\IEEEoverridecommandlockouts
\usepackage{cite}
\usepackage{amsmath,amssymb,amsfonts,booktabs,enumitem,multirow,array,algorithm,algpseudocode}
\usepackage{graphicx}
\usepackage{textcomp}
\usepackage{xcolor}
\usepackage{tikz}
\usetikzlibrary{arrows.meta, positioning}

\usepackage{amsthm}

\theoremstyle{definition}
\newtheorem{definition}{Definition}[section]
\newtheorem{example}{Example}

\newcommand{\NLQ}{\mathbf{NLQ}}
\newcommand{\Spans}{\mathbf{Spans}}
\newcommand{\T}{\mathbf{T}}
\newcommand{\C}{\mathbf{C}}
\newcommand{\V}{\mathbf{V}}
\newcommand{\Ents}{\mathcal{E}}

\def\BibTeX{{\rm B\kern-.05em{\sc i\kern-.025em b}\kern-.08em
    T\kern-.1667em\lower.7ex\hbox{E}\kern-.125emX}}

\begin{document}

\title{
    Database Entity Recognition with Data Augmentation and Deep Learning
}

\author{\IEEEauthorblockN{Zikun Fu}
\IEEEauthorblockA{\textit{Faculty of Science} \\
\textit{Ontario Tech University}\\
Oshawa, ON, Canada \\
zikun.fu@ontariotechu.net}
\and
\IEEEauthorblockN{Chen Yang}
\IEEEauthorblockA{\textit{College of Engineering} \\
\textit{Northeastern University}\\
Boston, MA, United States \\
yang.chen9@northeastern.edu
}
\and
\IEEEauthorblockN{Kourosh Davoudi}
\IEEEauthorblockA{\textit{Faculty of Science} \\
\textit{Ontario Tech University}\\
Oshawa, ON, Canada \\
heidar.davoudi@ontariotechu.ca}
\and
\IEEEauthorblockN{Ken Q. Pu}
\IEEEauthorblockA{\textit{Faculty of Science} \\
\textit{Ontario Tech University}\\
Oshawa, ON, Canada \\
ken.pu@ontariotechu.ca}
}

\maketitle

\begin{abstract}
This paper addresses the challenge of Database Entity Recognition (DB-ER) in Natural Language Queries (NLQ). We present several key contributions to advance this field: (1) a human-annotated benchmark for DB-ER task, derived from popular text-to-sql benchmarks \cite{bird, spider}, (2) a novel data augmentation procedure that leverages automatic annotation of NLQs based on the corresponding SQL queries which are available in popular text-to-SQL benchmarks, (3) a specialized language model based entity recognition model using T5 as a backbone and two down-stream DB-ER tasks: sequence tagging and token classification for fine-tuning of backend and performing DB-ER respectively.  We compared our DB-ER tagger with two state-of-the-art NER taggers, and observed {\em better} performance in both precision and recall for our model.  The ablation evaluation shows that data augmentation boosts precision and recall by over $10\%$, while fine-tuning of the T5 backbone boosts these metrics by $5-10\%$.
\end{abstract}


\section{Introduction}

Named Entity Recognition (NER) is a fundamental task in Natural Language Processing (NLP) that involves identifying and classifying named entities in text into predefined categories. While traditional NER focuses on general entity types, we address a more specialized challenge: database entity recognition in natural language queries (NLQ). This task is particularly crucial in the context of text-to-SQL systems, where accurately identifying database entities in natural language queries serves as a vital intermediate step. By properly tagging database entities in NLQs, downstream models can more effectively generate and validate complete queries across various query languages.

In this paper, we specifically tackle the Database Entity Recognition (DB-ER) problem using deep learning approaches. Our work distinguishes itself from conventional NER systems by leveraging existing text-to-SQL training data, where each NLQ is paired with its corresponding SQL query. This unique pairing provides an opportunity to enhance our understanding of database entities through the structural information present in SQL queries, enabling more sophisticated data augmentation techniques.

To facilitate the creation of high-quality training data for our DB-ER model (defined in Section~\ref{sec:model}), we have developed a web application that supports collaborative annotation of NLQ samples. This tool enables multiple users to concurrently annotate database entities while considering both the underlying database schema and the corresponding SQL query.  Figure~\ref{fig:labr} shows part of the user interface.

Given the inherent challenges of manual annotation, including high costs and limited efficiency, we introduce an innovative data augmentation algorithm. This algorithm automatically annotates NLQs by utilizing the structural information from their corresponding SQL queries. This approach significantly expands our training dataset while maintaining annotation quality, providing a scalable solution to the data scarcity problem in DB-ER due to the labour intensive nature of data labeling.

\begin{figure}
    \centering
    \includegraphics[width=1\linewidth]{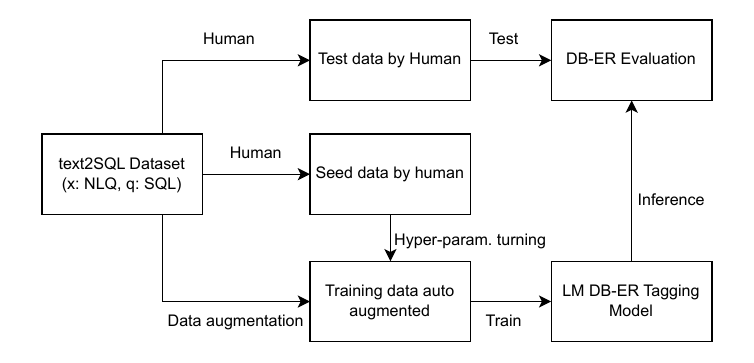}
    \caption{Data augmentation pipeline for training and validating DB-ER tagger. 
 In our pipeline, human generated annotations are augmented with auto-annotation. 
 The DB-ER tagger trained on the augmented dataset is then evaluated against the human annotated test dataset, which was held-out in the data augmentation process.}
    \label{fig:pipeline}
\end{figure}

Our technical contribution includes a specialized DB-ER tagger built upon the T5 language model architecture. Through extensive experimentation and ablation studies, we have identified two key techniques that substantially improve DB-ER performance: (1) data augmentation using our auto-annotation system, which boosts accuracy by over 10\%, and (2) fine-tuning the T5 backbone with a seq2seq learning task, which further improves end-to-end tagging accuracy by 5-10\%. These improvements demonstrate the effectiveness of our approach in addressing the challenges of database entity recognition.  Figure~\ref{fig:pipeline} shows the overall pipeline including data augmentation, training, and evaluation.

\section{Problem description}

In this section, we formally define the Database Entity Recognition (DB-ER) problem, and outline a machine learning approach to the problem and data augmentation with synthetic annotation.

\subsection{DB-ER tagging}

\begin{definition}[NLQ and Spans]
A {\em NLQ} $x$ is a sequence of words (or tokens). Given $x\in\NLQ$, we denote $\Spans(x)$ as all token subsequences of $x$.  Each {\em span} $s\in\Spans(x)$ is characterized by the start and end position: $(i,j)$ where $0\leq i\leq j< |x|$.
\end{definition}
In the context of database entity recognition, we need to identify specific spans within a NLQ that correspond to database entities. These entities can be of different types, as defined below.

\begin{definition}[Database Entity]
A {\em database entity} $e = (\mathrm{text}(e), \mathrm{type}(e))$ where
$\mathrm{text}(e)$ is the text value of the entity as a sequence of tokens, and
$\mathrm{type}(e)$ is the type of the entity, which can be one of the following:
    {\em table names} ($\T$), {\em column names} ($\C$), {\em cell values} ($\V$).    
We denote the set of all database entities\footnote{
In this paper, database entities $\Ents$ refer to the typed string values of table names, column names and database values.  This is not to be confused with entities in traditional Entity-Relation (ER) modeling.
}
as $\Ents$.  
\end{definition}

\begin{definition}[Annotation]
    An annotation $h_x$ of $x\in\NLQ$ is a partial mapping from $\Spans(x)$ to $\Ents$.
    Alternatively, we can think of $h_x$ as sequence labeling of tokens in $x$ by
    labels from $\Ents\cup\{\mathrm{none}\}$ where $\mathrm{none}$ means that the corresponding
    token is not part of a database entity.  We define the space of all possible DB-ER annotations as $\mathcal{A} = \{h_x \mid x \in \NLQ\}$.
\end{definition}

A DB-ER tagger is a function $H:\NLQ\to\mathcal{A}$.  We focus on a machine learning approach to DB-ER.  Namely, we will train a model $H_\theta$ as the DB-ER tagger that will take NLQ as input and output a DB-ER annotation $h_x$.  The architecture is given by a two-headed deep neural network in which we use T5-encoder has the embedding backbone, and have two separate heads: T5-decoder for sequence tagging, and a linear classifier for per-token classification into $\{\T, \C, \V, \mathrm{none}\}$.


\begin{center}
\begin{tikzpicture}[
    node distance=0.5cm and 1.8cm,
    every node/.style={font=\footnotesize},
    arrow/.style={->, thick}
]

\node (x)                             {$x$};
\node (z)         [right=of x]        {$z$};
\node (yseq)      [above right=of z]  {$y_{\mathrm{seq}}$};
\node (ycls)      [below right=of z]  {$y_{\mathrm{cls}}[i]$};

\draw[arrow] (x) -- node[above] {\footnotesize $\mathbf{T5Enc}$} (z);
\draw[arrow] (z) -- node[above left] {\footnotesize $\mathbf{T5Dec}$} (yseq);
\draw[arrow] (z) -- node[below left] {\footnotesize $\mathbf{Classifier}$} (ycls);

\end{tikzpicture}
\end{center}

\subsection{Synthetic annotations}

The training data for DB-ER comes from text-to-SQL datasets such as Spider \cite{spider} and BIRD \cite{bird}. These datasets contain pairs of natural language queries and their corresponding SQL queries. We leverage this pairing to help prepare training data for DB-ER that optimally matches spans of NLQ with database entities in the corresponding SQL query.  The synthetic annotation is a function:
$$
\mathbf{syn}: \NLQ\times\mathbf{SQL} \to\mathcal{A}
$$
The details of data augmentation is given in Section~\ref{sec:annotation}.
We combine the ground truth annotations and the automatically generated annotations to create a larger training set for the DB-ER model.

\begin{figure}[t]
    \centering
    \includegraphics[width=\linewidth]{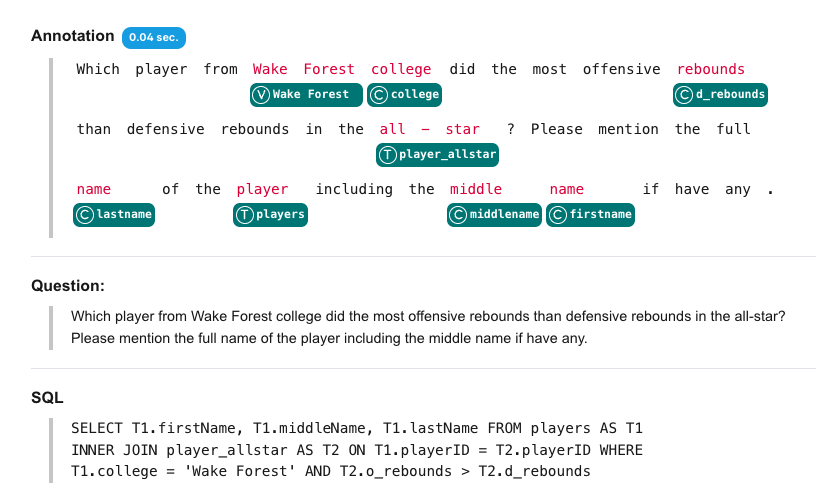}
    \caption{The collaborative annotation interface.}
    \label{fig:labr}
\end{figure}

\section{Related work}
\subsection{Named entity recognition}

\noindent\textbf{Traditional NER} have achieved significant success by leveraging contextual embeddings and deep learning architectures. LUKE~\cite{luke} introduces entity-aware self-attention mechanisms to improve entity representations, while Flair~\cite{flair} utilizes contextual string embeddings to capture character-level features. spaCy~\cite{spacy} offers an efficient pipeline for NLP tasks, including a robust NER model. These models achieve high accuracy in standard benchmarks (e.g., CoNLL-2003), but they are typically trained on open-domain entities (person, organization, location, etc.) and may falter on specialized domains with unseen entity vocabularies.\\

\subsection{Text-to-SQL and schema linking}

The task of text-to-SQL parsing has spurred the development of cross-domain benchmarks such as BIRD~\cite{bird} and Spider~\cite{spider}. 
A core challenge in text-to-SQL is schema linking \cite{schemalinking} – identifying which parts of the question correspond to which database tables or columns. Many text-to-SQL models explicitly incorporate schema linking modules or features~\cite{CDtext2sqlReview}, essentially performing a form of NER to tag tokens with the relevant database field. Our work can be seen as learning such a tagging function explicitly: by converting text-to-SQL data into a token-level entity labeling task, we provide a focused benchmark for evaluating and improving schema linking performance in isolation \cite{fu2024transforming}.

\begin{algorithm}[t]
\caption{Synthetic annotation}
\label{alg:syn}
\begin{algorithmic}[1]
\Require String similarity measure $\mathrm{sim}$, threshold $c\in\mathbb{R}^+$
%
\Function{Syn}{$x, sql, \mathrm{sim}, c$}
  \State $E \gets \text{Extract-DB-Entities}(sql)$ 
  \ForAll{$e \in E$} \Comment{Pair-wise similarity score}
    \ForAll{$\{s\in\Spans(x) \mid \mathrm{sim}(s, e)\geq c\}$}
    \State $\mathbf{SCORE}[e,s] \gets \mathrm{sim}(e,s)$
    \EndFor
  \EndFor
  \State // formulate an integer linear program (ILP)
  \State $\mathbf{ILP} \gets \mathrm{nonoverlap}(x, E, \mathbf{SCORES})$
  \State $\mathbf{OBJ} \gets \mathrm{objective}(\mathbf{SCORES})$
  \State $s^* \gets \mathrm{solve}(\mathbf{ILP}, \mathbf{OBJ})$ \Comment{optimal annotation}
  \State \Return $\{\,i \rightarrow e \mid i \in s^*(e)\,\}$
\EndFunction
\end{algorithmic}
\end{algorithm}

\begin{algorithm}[t]
    \caption{Data augmentation pipeline}
    \label{alg:aug}
    \begin{algorithmic}[1]
    \Require ${D}_\mathrm{hum} = \{(x_i, h_{xi})\}$: human annotated samples
    \Require ${D}_\mathrm{text2sql} = \{ (x_j, sql_j) \}$: raw text-to-sql samples
    \Function{DataAugment}{${D}_\mathrm{human}, {D}_\mathrm{text2sql}$}
        \State // perform grid search over hyperparameter space
        \State $X = \{\mathrm{jaccard},\mathrm{levenstein}\} \times [0.1, 0.2, \dots 1.0]$
        \State $(\mathrm{sim}^*, c^*) = \mathrm{argmax}(
            \mathbf{F}_1(
                D_\mathrm{hum},
                \mathrm{map}(\mathbf{Syn}, D_\mathrm{hum})
            )
        )$
        \State // augment data using optimal hyperparameter
        \State $D_\mathrm{augmented} \gets 
        \mathrm{map}(\mathrm{Syn}, D_\mathrm{text2sql}$
    \EndFunction
    \end{algorithmic}
\end{algorithm}

\section{Annotation and data augmentation}
\label{sec:annotation}

\subsection{Human annotation}

To facilitate the creation of high-quality training data for our DB-ER task, we implemented a web-based collaborative annotation platform. This platform enables multiple users to simultaneously annotate natural language queries (NLQs) while considering both the underlying database schema and the corresponding SQL queries. The collaborative nature of this platform ensures consistent and accurate annotations across the dataset.  The annotation interface, as shown in Figure~\ref{fig:labr}, provides an intuitive environment for human annotators.

This collaborative annotation system has been instrumental in creating a high-quality dataset for training our DB-ER model. The platform's ability to maintain consistency across annotations while considering both the NLQ and SQL query contexts has significantly improved the quality of our training data.

\subsection{Data augmentation with synthetic annotations}

Given the high cost of human annotation, we also explored data augmentation techniques to increase the size of our training data.  See Table~\ref{tab:split-stats} for the detailed data volumes of human vs augmented data.

Available from text-to-sql datasets are pairs of NLQs and their corresponding SQL queries: $(x, \mathrm{SQL}(x))$.  For the purpose of data augmentation, we want to devise a function $\mathrm{syn}: \NLQ\times\mathrm{SQL}\to\mathcal{A}$ such that output annotation $\mathrm{syn}(x, \mathrm{SQL}(x)) \in \mathcal{A}$ is a high-quality database entity annotation for the input NLQ $x$.

It is worth noting that $\mathrm{syn}$ cannot be used as a solution to the DB-ER task, since the input to $\mathrm{syn}$ is a NLQ and a SQL query, not an NLQ alone.  However, $\mathrm{syn}$ can be applied to the entire text-to-sql dataset because each NLQ in the dataset is paired with its corresponding SQL query.

{\em Step 1. DB Entity Identification.}  Let $q = \mathrm{SQL}(x)$.  We parse $q$ into its abstract syntax tree (AST), and traverse through the AST to extract all references to database tables, columns and values.  Let $E(q)\subseteq\Ents$ be the set of entities extracted from $q$.

{\em Step 2. Candidate spans generation.} Find matching spans in the NLQ for each entity in $E(q)$.  For each entity $e\in E(q)$, we find all spans of $x$ that are at least $c$-similar to $e$ according to the string similarity function $\mathrm{sim}$.
$$
S(x, e) = \{ s \in \Spans(x) \mid \mathrm{sim}(s, e) \geq c \}
$$
Here $\mathrm{sim}:\mathrm{string}\times\mathrm{string}\to\mathbb{R}^+$ and $c\in\mathbb{R}^+$ are hyperparameters that can be tuned for the data augmentation stage.  We have used two similarity measures: Jaccard similarity between 3-grams of two strings, and Levenstein edit distance between two strings \cite{vijaymeena2016survey}.  The threshold $c$ is selected during data augmentation validation set (see Step 4. below.).

{\em Step 3. Annotation generation.} In this step, we need to select zero or one span from each $s^*(e) \in S(x, e)$ such that $\{s^*(e) \mid e\in E(q)\}$ is a valid annotation for $x$.
The selection of $s^*(e)$ is based on the following criteria:
\begin{itemize}
    \item non-overlapping: if $e\not= e'$, then $s^*(e)\cap s^*(e') = \emptyset$.
    \item maximal matching: $\sum_{e\in E(q)} \sim(s^*(e), e)$ is maximized.
    \item optional selection: $s^*(e)$ is allowed to be empty for any $e\in E(q)$.
\end{itemize}
These constraints can be expressed as an integer linear program, and can be efficiently solved optimally using an integer programming solver \cite{cpsatlp}.

Algorithm~\ref{alg:syn} shows the high level pseudo code for computing the optimal sequence label $s^*$, and $\mathrm{syn}(x, q) = \{i \to e\in E(q) \mid i\in s^*(e), i\in \mathrm{tokens}(x)\}$.

{\em Step 4. Data augmentation validation.}  Data augmentation can be validated using the human annotated data.  We compare the synthetic annotated database entity with human annotated entity at the token level using the standard information retrieval F1-measure.  The validation step allows us to perform grid search to select the optimal threshold $c$ and the string similarity measure.  The optimal similarity measure and its threshold is used to generate the final augmented training data from text-to-sql dataset.  The pseudo code for data augmentation is shown in Algorithm~\ref{alg:aug}.

\section{T5 based DB-ER tagging: model and training}
\label{sec:model}

\begin{figure}
    \centering
    \includegraphics[width=1\linewidth]{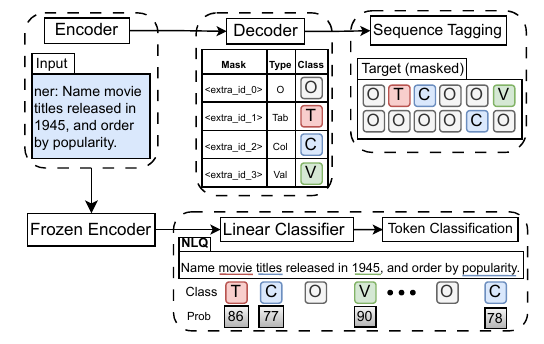}
    \caption{Two-stage pipeline for database-entity tagging. 
    (a) T5 is fine-tuned by masked infilling so its encoder learns schema-aware representations. (b) The encoder is frozen and a lightweight linear classifier is trained on the frozen token embeddings with human labels to output \textsc{TABLE}, \textsc{COLUMN}, \textsc{VALUE}, or \textsc{O}.}
    \label{fig:T5_arch}
\end{figure}

Our tagging pipeline (Fig.~\ref{fig:T5_arch}) follows a two–stage design.  
\emph{Stage 1} (\emph{Sequence Tagging}) fine-tunes a T5 encoder–decoder \cite{raffel2020exploring} so that the encoder learns schema-aware representations by reconstructing a sequence of entity masks (\textsc{TABLE}, \textsc{COLUMN}, \textsc{VALUE}, \textsc{O}) from the input question.  
\emph{Stage 2} (\emph{Token Classification}) freezes the encoder and trains a lightweight linear layer that maps each contextual embedding to its final label.

Directly decoding tags with T5 yields an F\textsubscript{1} of only \textbf{0.40} on the human test set (Fig.~5). By decoupling representation learning from classification and fitting a dedicated linear head, the same encoder attains $\textbf{0.77}$ in F1.  

\subsection{T5 fine-tuning with sequence tagging}
\label{sec:t5_seqtag}

We used a T5 sequence-to-sequence language model as the backbone for database entity tagging. The model is fine-tuned to predict entity labels for each token in an input question. To achieve this, we perform sequence-to-sequence learning to generate the output tag sequence.  We call this task {\em sequence tagging}.

During fine-tuning, the NLQ is passed to the encoder, and the decoder is trained to output a sequence of tag identifiers (using the reserved \texttt{extra} tokens). This allows the model to learn entity-aware generation conditioned on natural language input. We fine-tune both T5-Base (220M) and T5-Large (770M) on two data sources: (1) a small set of \textbf{human-annotated} examples, and (2) a large-scale \textbf{synthetic dataset} derived from auto-tagging BIRD and Spider benchmark questions.

The goal of this fine-tuning stage is to adapt the encoder representations to the database domain, such that entity-relevant semantics are encoded in token embeddings. Fine-tuning is run until convergence using cross-entropy loss on the predicted tag sequences: 
\texttt{<id\_0>} for \texttt{none}, 
\texttt{<id\_1>} for \textbf{T},
\texttt{<id\_2>} for \textbf{C},
\texttt{<id\_3>} for \textbf{V}.

\begin{example}
Consider the NLQ: $x =${\em ``Name movie titles released in 1945, and order by popularity''}.
The SQL query is given by:
\begin{verbatim}
  SELECT title FROM movies 
  WHERE year = 1945 ORDER BY pop
\end{verbatim}
The extracted database entities $E$ from sql are:
$$
E = \{\mathrm{title}:\mathbf{C},
\mathrm{movies}:\mathbf{T},
\mathrm{year}:\mathbf{C},
\mathrm{1945}:\mathbf{V},
\mathrm{pop}:\mathbf{C}\}
$$
The synthetic annotation based on Jaccard similarity measures would produce
the following (abbreviated) annotation:
$$
\mathrm{Name}
\ \underbrace{\mathrm{movie}}_{\mathrm{movies}:\mathbf{T}}
\ \underbrace{\mathrm{titles}}_{\mathrm{title}:\mathbf{C}}
\dots \mathrm{\ order\ by\ } \underbrace{\mathrm{popularity}}_{\mathrm{pop}:\mathbf{C}}.
$$
The sequence tagging output is the entity type labels of each of the NLQ tokens.
$$
\underbrace{\mathrm{Name}}_{N}
\ \underbrace{\mathrm{movie}}_{\mathbf{T}}
\ \underbrace{\mathrm{titles}}_{\mathbf{C}}
\dots 
\underbrace{\mathrm{order}}_{N}
\ \underbrace{\mathrm{by}}_{N}
\ \underbrace{\mathrm{popularity}}_{\mathbf{C}}.
$$
This would be one example used to fine-tune the T5 model.
\qed
\end{example}

\subsection{Token classification using a linear classifier}
\label{sec:linear_cls}

We have observed that sequence tagging is a data-intensive learning task, requiring much more training data to reach high accuracy (see Figure~\ref{fig:t5-f1-comparison}).  With this observation in mind, we introduce a classification head that performs token classification for each token embedding produced by the T5-encoder.  As our experimental evaluation demonstrates, token classification requires less data due to the simplicity of the classification head.

Once T5 is fine-tuned, we freeze its encoder and extract contextual token embeddings for each NLQ/SQL pair. A lightweight linear classifier is then trained to assign each token an entity class: \textsc{Table}, \textsc{Column}, \textsc{Value}, or \textsc{O} (non-entity). The classifier consists of a single fully connected layer followed by softmax.

\begin{table}[ht]
    \centering
    \caption{Data splits Statistics}
    \label{tab:split-stats}
    \begin{tabular}{lrrrrrr}
    \toprule
    \multirow{2}{*}{Entity} & \multicolumn{2}{c}{\textbf{Human Train}} & \multicolumn{2}{c}{\textbf{Synthetic Train}} & \multicolumn{2}{c}{\textbf{Human Test}}\\
        \cmidrule(lr){2-3}\cmidrule(lr){4-5}\cmidrule(lr){6-7}
         & Tokens & \% & Tokens & \% & Tokens & \%\\
    \midrule
        \textsc{Table}   &   592 & 7.9  & 17\,874 & 7.8  &   616 & 8.3 \\
        \textsc{Column}  & 1\,018 & 13.6 & 31\,023 & 13.6 & 1\,105 & 14.9\\
        \textsc{Value}   &   563 & 7.5  & 18\,413 & 8.0  &   517 & 7.0 \\
        \textsc{O}       & 5\,303 & 70.9 & 161\,475 & 70.6 & 5\,175 & 69.8\\
        \midrule
        \textbf{Total}   & 7\,476 & 100  & 228\,785 & 100  & 7\,413 & 100\\
    \bottomrule
    \end{tabular}
\end{table}

\begin{table*}[ht]
  \centering
  \caption{Token-level precision (P), recall (R), and F1 (\%) on the human test set.\\
  Abbreviations: 
  pre-trained (PT), finetuned (FT), Human annotated (H), Augmented with synthetic annotation (S)}
  \label{tab:main-results}
  \resizebox{\textwidth}{!}{\tiny
  \begin{tabular}{lccccccccccccc}
    \toprule
    \textbf{Model} & \textbf{Configuration} &
      \multicolumn{3}{c}{\textsc{Table}} &
      \multicolumn{3}{c}{\textsc{Column}} &
      \multicolumn{3}{c}{\textsc{Value}} &
      \multicolumn{3}{c}{\textsc{I} (2-class)}\\
     & & Prec & Recall & F1 & Prec & Recall & F1 & Prec & Recall & F1 & Prec & Recall & F1\\
    \cmidrule(lr){3-5}\cmidrule(lr){6-8}\cmidrule(lr){9-11}\cmidrule(lr){12-14}
    \multicolumn{14}{c}{\textit{SOTA NER models}}\\
    \midrule
    \multirow{3}{*}{LUKE (large)}
      & PT/H & -- & -- & -- & -- & -- & -- & -- & -- & -- & 22.9 & 5.4 & 8.7\\
      & FT/H & 52.7 & 53.5 & 53.1 & 55.3 & 30.9 & 39.6 & 55.1 & 54.8 & 55.0 & 65.3 & 51.9 & 57.9\\
      & FT/S & 70.3 & 61.6 & 65.7 & 63.9 & 44.8 & 52.7 & 72.3 & 75.6 & 73.9 & 76.0 & 63.0 & 68.9\\
    \midrule
    \multirow{3}{*}{Flair (large)}
      & PT/H & -- & -- & -- & -- & -- & -- & -- & -- & -- & 81.4 & 10.8 & 19.0\\
      & FT/H & -- & -- & -- & -- & -- & -- & -- & -- & -- & 77.3 & 73.4 & 75.3\\
      & FT/S & -- & -- & -- & -- & -- & -- & -- & -- & -- & 80.5 & 78.0 & 79.3\\
    \midrule
    \multicolumn{14}{c}{\textit{T5-based models (ours)}}\\
    \midrule
    \multirow{4}{*}{T5 (base)}
    & PT/H & 55.5 & 48.8 & 51.9 & 63.1 & 47.8 & 54.4 & 65.0 & 70.2 & 67.5 & 79.9 & 69.0 & 74.1\\
    & PT/S & 62.0 & 60.7 & 61.4 & 65.5 & 56.4 & 60.6 & 67.1 & 76.5 & 71.5 & 80.9 & 77.4 & 79.1\\
    & FT/H & 54.1 & 48.6 & 51.2 & 63.0 & 48.2 & 54.6 & 66.1 & 70.6 & 68.3 & 79.6 & 69.4 & 74.1\\
    & FT/S & 70.4 & \textbf{67.3} & 68.8 & 69.2 & 62.3 & 65.5 & 75.8 & \textbf{80.7} & \textbf{78.2} & 82.9 & \textbf{79.0} & \textbf{80.9}\\
    \midrule
    \multirow{4}{*}{T5 (large)}
    & PT/H & 58.8 & 51.2 & 54.8 & 65.5 & 49.9 & 56.6 & 65.5 & 67.6 & 66.5 & 80.3 & 68.5 & 73.9\\
    & PT/S & 63.5 & 62.4 & 62.9 & 64.4 & 53.7 & 58.6 & 66.5 & 74.0 & 70.1 & 80.4 & 75.4 & 77.8\\
    & FT/H & 54.2 & 47.8 & 50.8 & 62.2 & 50.4 & 55.7 & 65.1 & 65.4 & 65.3 & 79.6 & 69.5 & 74.2\\
    & FT/S & \textbf{73.6} & 67.1 & \textbf{70.2} & \textbf{72.2} & \textbf{65.7} & \textbf{68.8} & \textbf{76.9} & 78.9 & 77.9 & \textbf{83.5} & 78.2 & 80.8\\
    \bottomrule
  \end{tabular}}
\end{table*}

\begin{table}[ht]
\centering
\small
\caption{T5 DB-NER tagger architecture configurations}
\label{tab:configs}
\renewcommand{\arraystretch}{1.5} 
\begin{tabular}{|c|p{0.7\linewidth}|} \hline
{\bf Configuration} & {\bf Architecture} \\ \hline
{\tt PT/H} & {\em Pre-trained} T5 encoder + classifier trained on data with {\em human} annotation. \\ \hline
{\tt PT/S} & {\em Pre-trained} T5 encoder + classifier trained on augmented data with {\em synthetic} annotation. \\ \hline
{\tt FT/H} & {\em Fine-tuned} T5 encoder + classifier on data with {\em human} annotation. \\ \hline
{\tt FT/S} & {\em Fine-tuned} T5 encoder + classifier on augmented data with {\em synthetic} annotation. \\ \hline
\end{tabular}
\end{table}
\section{Training and evaluation}
\label{sec:training-eval}


We benchmark three families of taggers:

\begin{enumerate}
    \item \textbf{SOTA baselines}: LUKE-Large\footnote{huggingface.co/studio-ousia/luke-large} and Flair-Large\footnote{huggingface.co/flair/ner-english-large}.  LUKE is pre-trained on English Wikipedia and fine-tuned with our data; Flair is trained on CoNLL-2003, and finetuned with our data using the official flair-NER recipe.
    \item \textbf{Sequence Tagging}: T5 encoder–decoder predicts tag identifiers.
    \item \textbf{Token Classification}: the same frozen T5 encoder with a linear head.
\end{enumerate}

\subsection{Dataset splits}

Dataset statistics are shown in Table~\ref{tab:split-stats}. All three splits exhibit a near-identical distribution—about \textbf{29\%} entity tokens vs.\ \textbf{71\%} background \textsc{O}. The pronounced class imbalance motivates the use both fine-grained \emph{4-class} metrics and a coarse \textbf{entity-vs-non-entity (2-I)} score that is less sensitive to the prevalence of \textsc{O}.

All models are evaluated on the same human test set comprising 500 manually labeled examples. Training sets include:
\begin{itemize}
    \item \textbf{Human-train}: 1,000 human-annotated examples (gold labels).
    \item \textbf{Synthetic-train}: 15,000 examples auto-tagged using our pipeline from the BIRD and Spider datasets.
\end{itemize}

For each T5 size (\textbf{Base} and \textbf{Large}), we explore four configurations based on the type of training and data used as shown in Table~\ref{tab:configs}. In the PT/H setting, the pre-trained T5 encoder is paired with a classifier trained on human-annotated data. PT/S uses the same pre-trained encoder, but the classifier is trained on synthetically annotated data. In FT/H, the encoder itself is fine-tuned alongside the classifier using human-annotated data. Lastly, FT/S fine-tunes the encoder and trains the classifier on synthetic annotations.

\subsection{Evaluation metrics}

Token-level precision, recall, and F\textsubscript{1} are computed for

\begin{itemize}
    \item \textbf{4-class}: Table, Column, Value, O
    \item \textbf{3-class}: S = (Table + Column), V = Value, O
    \item \textbf{2-class}: I = any entity (Table, Column, Value) vs. O
\end{itemize}

All metrics are computed by aligning predicted labels with ground-truth annotations on the 500-example human test set.

\subsection{Evaluation results}

\begin{figure}[ht]
  \centering
  \includegraphics[width=0.8\linewidth]{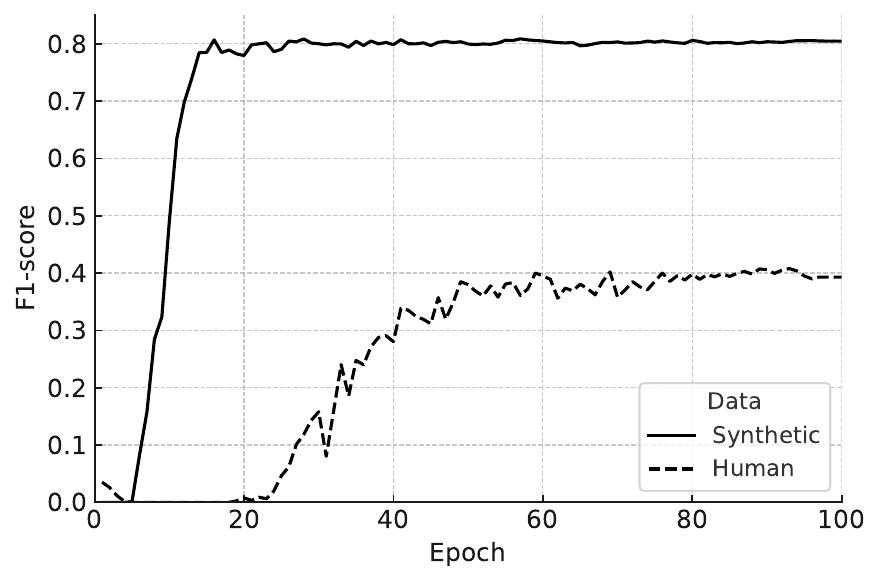}
  \caption{
  $F_1$ score for our model with T5 (large) during training over 100 epochs. Here we compare the in-training $F_1$ score in two scenarios: (1) using human annotations only and (2) using augmented data with synthetic annotations.
  }
  \label{fig:entity-f1}
\end{figure}

\begin{figure}[t]
  \centering
  \includegraphics[width=0.8\linewidth]{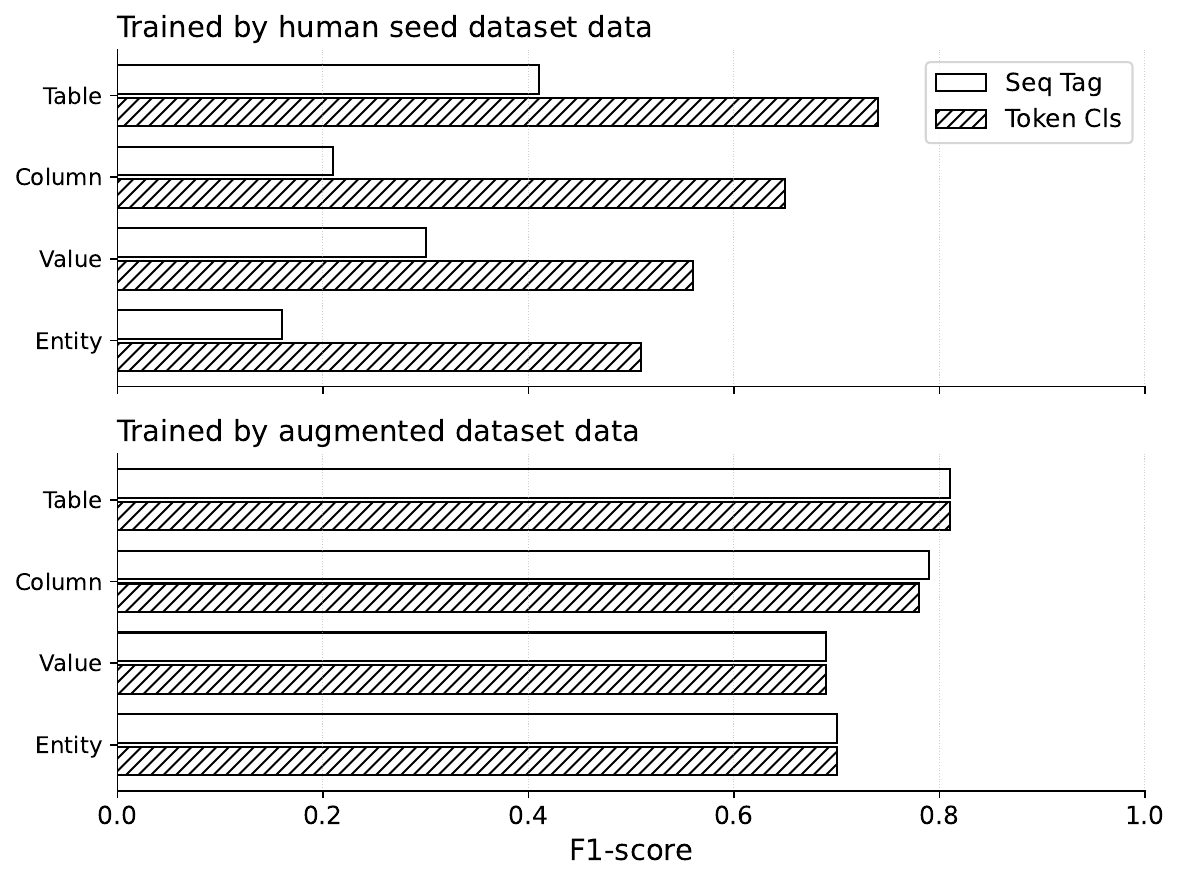}
  \caption{
    $F_1$ scores of our DB-ER tagger with T5 (large) as backbone on the four DB entity categories. Bars grouped by category show: (i) \emph{Sequence Tagging} model with T5 decoder; and (ii) two-stage \emph{Token Classification} pipeline (frozen encoder with linear classifier). The classifier yields large gains on scarce human data while both approaches converge when abundant synthetic supervision is available.}
  \label{fig:t5-f1-comparison}
\end{figure}

Table~\ref{tab:main-results} presents the final numbers; Fig.~\ref{fig:entity-f1} tracks T5-\textbf{large} during training; Fig.~\ref{fig:t5-f1-comparison} contrasts the two heads.

\paragraph*{Synthetic supervision gives better results}
Across all backbones and heads, \texttt{FT/S} outperforms \texttt{FT/H} by 14–19 F\textsubscript{1} on Tables, 10–15 on Columns, and 10–12 on Values.  
For T5-Large the jump in overall entity F\textsubscript{1} (2-I) is \textbf{+9.3} (0.69\,$\to$\,0.78).  
Fig.~\ref{fig:entity-f1} shows the gap opening within the first 20 epochs and persisting thereafter.

\paragraph*{Token Classification beats Sequence Tagging on scarce data.}
With only 1k human examples, decoding tags with the T5 decoder tops out at 0.40 I-F\textsubscript{1}; adding a linear head on the same encoder reaches 0.77 (Fig.~\ref{fig:t5-f1-comparison}).  
The advantage narrows when 15k synthetic examples are available.

\paragraph*{T5 vs.\ SOTA baselines}
The best LUKE configuration (\texttt{FT/S}) hits 0.689 I-F\textsubscript{1}, trailing T5-Large by 11.9 pts.  
Flair’s off-the-shelf embeddings fail on our schema vocabulary (0.19 I-F\textsubscript{1}), but fine-tuning raises performance to 0.79.

\paragraph*{Flair lacks 4-class scores}
Flair's pre-trained embedding model is trained on the CoNLL-2003 corpus, which includes only standard open-domain entities: \textsc{PER}, \textsc{LOC}, \textsc{ORG}, and \textsc{MISC}.  
Even when fine-tuned on our human or synthetic datasets, Flair fails to adapt to the schema-specific vocabulary required for \textsc{Table}, \textsc{Column}, or \textsc{Value} tagging.

\section{Conclusion and future work}

\subsection{Conclusion}
In this paper, we studied the problem of database entity recognition (DB-ER) in natural language queries and made several original contributions. First, we created a benchmark dataset by leveraging text-to-sql datasets \cite{bird,spider} and implementing a collaborative annotation platform that enables human annotators to accurately label database entities while considering both the database schema and SQL queries. Second, we introduced an novel data augmentation algorithm that automatically generates DB-ER annotations by analyzing the relationship between NLQs and their corresponding SQL queries, significantly expanding our training dataset while maintaining annotation quality.  

We constructed a specialized DB-ER tagger with a T5-backbone and a linear classifier. We developed a two-stage training strategy involving two tasks: sequence tagging and token classification tasks, allowing us to effectively fine-tune the backbone model and train a dedicated classifier. The experimental evaluation demonstrates the effectiveness of our data augmentation approach, showing significant improvements in model performance compared to previous NER taggers applied to our dataset. Specifically, our T5-based model achieves superior results across all entity types (tables, columns, and values), with particularly strong performance on table and column recognition tasks.

\subsection{Future work}

We are investigating the integration of our DB-ER tagger with downstream text-to-SQL systems to improve end-to-end query generation. Second, we plan to extend our data augmentation approach to handle more complex database schemas and query patterns. Finally, we could explore the application of our techniques to other database-related NLP tasks, such as schema-aware question answering or query understanding.

\bibliographystyle{plain}
\bibliography{references}
\end{document}